\documentclass{article}

\usepackage{mathtools}
\usepackage{microtype}
\usepackage{graphicx}
\usepackage{subcaption}
\usepackage{booktabs} % for professional tables
\usepackage{hyperref} 

 \usepackage[accepted]{icml2021}
\usepackage{amssymb,amsmath,amsthm} 
\usepackage{bm}
\usepackage{csvsimple}
\usepackage{pgfplotstable}
\graphicspath{{figs/}}

\usepackage[normalem]{ulem}

\DeclarePairedDelimiter\ceil{\lceil}{\rceil}

\newcommand{\vect}[1]{\boldsymbol{#1}}
\renewcommand{\AA}{\mathcal{A}}
\newcommand{\BB}{\mathcal{B}}

\newcommand{\DD}{\mathcal{D}}

\newcommand{\FF}{\mathcal{F}}

\newcommand{\LL}{\mathcal{L}} 
\newcommand{\MM}{\mathcal{M}}
\newcommand{\NN}{\mathcal{N}}

\newcommand{\UU}{\mathcal{U}}

\newcommand{\E}{\mathbb{E}}
\newcommand{\Var}{\mathbb{V}\text{ar}}

\newcommand{\m}{\mathop{\textrm{minimise}}}

\newcommand{\R}{\mathbb{R}}

\newcommand{\ie}{\emph{i.e.}}
\newcommand{\eg}{\emph{e.g.}}
\newcommand\oprocendsymbol{\hbox{$\square$}}
\newcommand\oprocend{\relax\ifmmode\else\unskip\hfill\fi\oprocendsymbol}

\theoremstyle{plain}

\theoremstyle{definition}

\theoremstyle{remark}

\icmltitlerunning{Intrinsic uncertainties and where to find them}

\begin{document}

\twocolumn[ \icmltitle{Intrinsic uncertainties and where to find them}

\icmlsetsymbol{equal}{*}

\begin{icmlauthorlist}
\icmlauthor{Francesco Farina}{to}
\icmlauthor{Lawrence Phillips}{to}
\icmlauthor{Nicola J Richmond}{to}
\end{icmlauthorlist}

\icmlaffiliation{to}{GSK.ai, GlaxoSmithKline, London, United Kingdom}

\icmlcorrespondingauthor{Francesco Farina}{francesco.x.farina@gsk.com}

\icmlkeywords{Machine Learning, ICML, Uncertainty estimation} 

\vskip 0.3in
]

% this must go after the closing bracket ] following \twocolumn[ ...

% This command actually creates the footnote in the first column
% listing the affiliations and the copyright notice.  The command
% takes one argument, which is text to display at the start of the
% footnote.  The \icmlEqualContribution command is standard text for
% equal contribution.  Remove it (just {}) if you do not need this
% facility.

\printAffiliationsAndNotice{}  % leave blank if no need to mention
                               % equal contribution
% \printAffiliationsAndNotice{\icmlEqualContribution} % otherwise use the standard text.

\begin{abstract}
We introduce a framework for
uncertainty estimation that both describes and extends many
existing methods. We consider typical hyperparameters involved
in classical training as random variables and marginalise them out to capture various sources of
uncertainty in the parameter space. 
We investigate which forms and combinations of marginalisation are most useful from a practical point of view
on standard benchmarking data sets. Moreover, we discuss how some marginalisations may produce reliable
estimates of uncertainty without the need for extensive hyperparameter tuning and/or large-scale ensembling.
\end{abstract}

\section{Introduction}
While modern neural network-based machine learning architectures have
yielded a step change in predicition accuracy over traditional
methods, especially in application domains such as computer vision,
speech recognition and natural language processing, accurately
estimating uncertainty remains an issue. Providing accurate
estimations of the uncertainty associated to each prediction is of
critical importance, particularly in the healthcare domain. Classical
Bayesian inference methods~\cite{wang2020survey,mackay1992bayesian,neal2012bayesian,murphy2012machine} address the uncertainty question but they
typically do not scale with architecture and data set size and so
resort to leveraging computationally-tractable approximations via
parametrisation.

This paper introduces a framework for
uncertainty estimation that provides a novel and simple way to describe 
and extend many uncertainty estimation methods.
The main idea relies on considering
hyperparameters involved in classical training procedures as random
variables. We show that through marginalising out different
combinations of hyperparameters, different sources of uncertainty
can be estimated in the (model) parameter space. Within this
framework, methods such as SWAG ~\cite{maddox_simple_2019},
deep ensembles ~\cite{lakshminarayanan2017simple}, MultiSWAG~\cite{wilson_bayesian_2020}, MC-dropout
~\cite{gal2016dropout} and hyperparameter ensembles
~\cite{levesque2016bayesian,wenzel2020hyperparameter} are shown to
be approximations to particular marginalisations.

We apply methods resulting from this framework on benchmarking
problems to investigate which forms and combinations of marginalisation 
are most useful from a practical point of view. 
As one may expect, results suggest
that increasing the number of random variables that are marginalised
out tends to increase the quality of the uncertainty
estimates, with certain random variables playing a more evident role. 
In particular, some combinations of marginalisations can produce reliable 
estimates of uncertainty without the need for extensive hyperparameter tuning. 
However, this is not the case with all combinations.

\section{Finding uncertainties}
Given a set of data points $\DD=\{x_i,y_i\}_{i=1}^N$,
one usually wishes to determine the predictive distribution $p(y\mid x,\DD)$ for some new value
$x$. As this problem is generally intractable, it is common practice
to assume the predictive distribution takes on a certain
functional form described by a parametric model $\FF_{\Theta} =
\{f_{\theta}:\R^n\to\R^m, \theta\in\Theta\}$ for some parameter space
$\Theta\subset\R^d$, and to determine 
\begin{equation}\label{eq:prediction}
  p(y\mid x, \DD, \FF_\Theta) = \int_\theta p(y\mid x,
    \theta) p(\theta\mid
  \DD)\mathrm{d}\theta.
\end{equation}
where the value of $y$, given $x$ and $\theta$ (and consequently $f_\theta$) is
deterministic and $p(\theta\mid \DD)$ is the distribution over $\theta$ that likely gave
rise to $\DD$.

\begin{figure*}
    \centering
    \includegraphics[width=\textwidth]{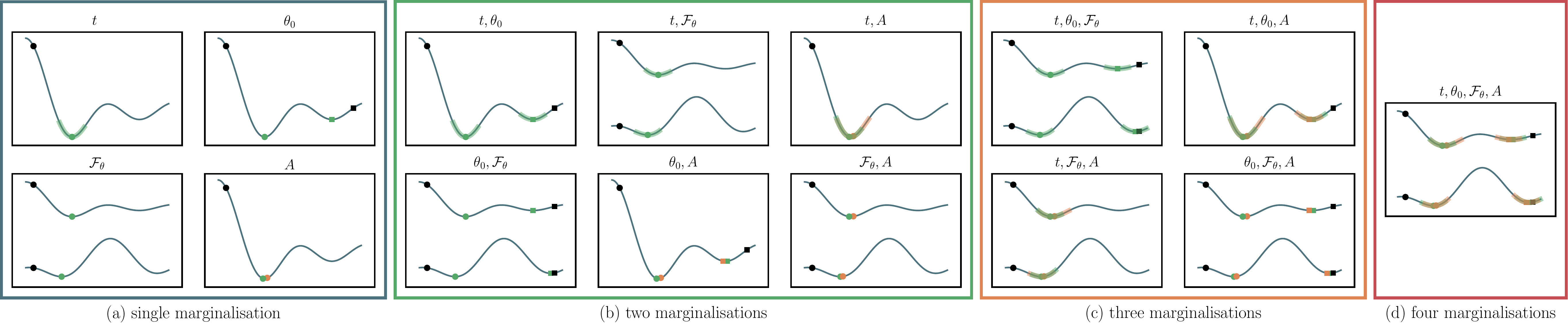}
    \caption{Pictorial view of the uncertainty induced by the different hyperparameters. The blue line represent the loss function $\LL$, black dots are the initial conditions $\theta_0$ and green and orange dots are (sub)optimal solutions after $t$ iterations of the algorithm. Shaded regions represent the possible position of $\theta_{t}$ for $t\geq\bar{\tau}$.}
    \label{fig:combinations}
\end{figure*}

For modern architectures and data sets, computing $p(\theta\mid \DD)$
is computationally intractable. This has given rise to a number of
proposed approximate methods. In the next sections, we show that many of these methods can be realised within our framework as
approximating the conditional probability distribution of $\theta$ given
the data set $\DD$ and additional random variables that capture
different aspects of the distribution, a graphical representation of
which is depicted in Figure~\ref{fig:combinations}.
\subsection{A starting point: classical training}
In the classical (deterministic) modern
neural network training paradigm, the training goal is to find the optimal model parameter
$\theta_\star\in\Theta$ that minimises some notion of difference
between $y_i$ and $f_\theta(x_i)$, usually referred to as loss
$\LL:\R^m\times \R^m\to\R$, by \emph{solving}
\begin{equation}\label{pb}
    \begin{aligned}
        &\m_\theta 
        & & \sum_{(x_i,y_i)\in\DD}\LL(y_i, f_\theta(x_i)).
    \end{aligned}
\end{equation}
Given $\theta_\star$, classical training methods can be seen as approximating $p(\theta\mid \DD)\approx\delta(\theta-\theta_\star\mid\DD)$.
Determining an exact solution to~\eqref{pb} is computationally
intractable for most problems arising in machine learning due to
factors such as data dimensionality and size, non-convexity of the
problem, etc. The aim then is to find a suboptimal solution via
an iterative algorithm $A_h$ from the countably infinite set of 
optimisation algorithms $\AA$, with chosen
hyperparameter point $h$ in an algorithm-appropriate hyperparameter
space $H$\footnote{Hereafter, when talking about hyperparameters we will refer only to those characterising the algorithm $A$.}.

Let $\theta_\tau$ be the solution estimate at iteration $\tau$ of
$A_h$. 
The sequence $\{\theta_\tau\}_{\tau=1}^t$
produced by $t$ iterations of $A_h$ is determined by the initial condition
$\theta_0\in\Theta_0\subset\R^d$ and the hyperparameters $h$, \eg, step-size, batch-size
and batch process order. With a slight abuse of notation, we consider
$A_h$ as a function $A_h:\R^d\to\R^d$ and write $\theta_t$ as
$A^t_h(\theta_0)$ to represent $t$ compositions (iterations) of
$A_h$.

It should now be clear that since problem~\eqref{pb} is generally
intractable, rather than simply making the approximation $p(\theta\mid \DD)\approx\delta(\theta-\theta_\star\mid\DD)$,
classical training computes the value of $p(\theta\mid \DD)$ further
conditioned on $\theta_0$, $h$ and $t$, \ie,
\begin{equation}\label{eq:classic}
    p(\theta\mid \DD,t,h,\theta_0)=\delta(\theta-A^t_h(\theta_0)\mid\DD).
\end{equation}
Then the output of a trained model $f_{\theta_t}$ is characterised as
\begin{align}
    p(y\mid x,\DD,t,h,\theta_0) 
    &=  \delta(y-f_{A^t_h(\theta_0)}(x)\mid\DD).\label{eq:T}
\end{align}

\subsection{Intrinsic uncertainties in classical training}

By marginalising out the different variables $\theta_0, t, h$ and possibly the model family $\FF$, we
should obtain estimates for $p(y\mid x, \DD)$ that are better than the point mass distribution~\eqref{eq:T}. 
In the next
sections, we will show that this is indeed the case and that many of
the current state-of-the-art algorithms for uncertainty estimation can
be considered as estimating $p(\theta\mid \DD)$ (or $p(y\mid x, \DD)$)
by marginalising out different subsets of the conditioning variables
in~\eqref{eq:classic} (or~\eqref{eq:T}).

\subsubsection{Number of iterations $t$}

Under suitable assumptions, algorithms in $\AA$ usually generate a
sequence of solution estimates that, in the limit, converges to a
neighbourhood of a stationary point $\hat{\theta}_\star$.
In practice, convergence to this neighbourhood usually occurs after a
certain number of iterations $\bar{\tau}$.
While in general there are no guarantees regarding the position of
$\theta_t$ in relation to $\hat{\theta}_\star$ within
the neighbourhood, the trajectory of the
solution estimates $\{\theta_\tau\}$ generated by the algorithm
provides information on the geometry of parameters space near $\hat{\theta}_\star$ which in turn
induces a probability distribution over $\Theta$ (see
Figure~\ref{fig:combinations}, plot $t$ in box (a)). Formally, we have
\begin{align*}
    p(\theta\mid \DD,h,\theta_0) = \lim_{t\to\infty}
    \frac{1}{t-\bar{\tau}}\sum_{\tau=\bar{\tau}}^t{\delta(\theta-A^t_h(\theta_0)\mid\DD)},
\end{align*}
which for stochastic gradient descent algorithms can be shown to be Gaussian in the
limit. Note that this formalism nicely describes the SWAG
algorithm~\cite{maddox_simple_2019}, which assumes $p(\theta\mid \DD,
h, \theta_0)$ is a Gaussian distribution and computes the first and
second order momenta by taking finite samples from
$\{\tau,\tau+1,\ldots\}$, and the TRADI
algorithm~\cite{franchi2019tradi} which again makes the Gaussian
assumption and tracks the momenta of $\Theta$ along the algorithm
evolution through a Kalman filter.

\subsubsection{Initial condition $\theta_0$}

Due to the likely non-convexity of the loss function, the initial
condition $\theta_0$ may determine the stationary point around which
$\{\theta_\tau\}$ stabilises. 
Thus, the initial
condition induces a probability distribution over $\Theta$ that can be
computed by marginalising~\eqref{eq:classic} with respect to
$\theta_0$, as
\begin{equation}\label{eq:t0}
    p(\theta\mid \DD,t,h) =
    \int_{\theta_0}\delta(\theta-A^t_h(\theta_0)\mid\DD)p(\theta_0)\mathrm{d}\theta_0,
\end{equation}
where $p(\theta_0)$ is chosen according to an appropriate parameter
initialisation strategy for the model architecture such as the Glorot
uniform distribution~\cite{pmlr-v9-glorot10a}. See
Figure~\ref{fig:combinations}, plot $\theta_0$ in box (a) for a
graphical representation.

Approximating~\eqref{eq:t0} is clearly the goal of classical ensemble
methods for uncertainty estimation (see,
\eg,~\citet{lakshminarayanan2017simple}).

\subsubsection{Algorithm $A$}

As mentioned, in general, different algorithm procedures and/or
different points in appropriate hyperparameter space produce different
approximate solutions to problem~\eqref{pb}. For example, in the case
of the simple Stochastic Gradient Descent (SGD), the step size, batch
size and the order in which batches are processed by the algorithm all
influence the trajectory $\{\theta_\tau\}$. This is also true for more
complex algorithms where other hyperparameters may play a role. To
capture the uncertainty associated to the choice of algorithm, we can
marginalise out $h$ from $p(\theta\mid
\DD,t,h,\theta_0)$ to give
\begin{equation*}%\label{eq:A}
    p(\theta\mid \DD,t,\theta_0) =
    \int_h\delta(\theta-A^t_h(\theta_0)\mid\DD)p(h)\mathrm{d}h.
\end{equation*}
In Appendix~\ref{sec:algos} we characterise for the SGD and for the general class of stochastic algorithms.
A graphical representation is reported in
Figure~\ref{fig:combinations}, plot $A$ in box (a), where different
choices of $h$ lead to different values of $A^t_h(\theta_0)$.

Ensembling over hyperparameter space to improve model performance and 
uncertainty estimation has been proposed in~\cite{levesque2016bayesian} 
and~\cite{wenzel2020hyperparameter}.

\subsubsection{Model family $\FF$}
Recall that assumptions are made about the functional form of the
predictive distribution to enable its computation. With this
in mind, so far, we have presented our framework in the
context of a fixed parametric model family $\FF_\Theta$ over some
parameter space $\Theta$. The particular choice of model family
typically involves architecture and hyperparameter space selection and
is usually informed by the problem domain and the available data set
$\DD$. As different parametric model families yield different
estimates of the predictive distribution, the choice of
model family is also a source of uncertainty which one can account for by
marginalising out $\FF_\Theta$ to give
\begin{equation}
    p(y\mid x, \DD) =\int_{\FF_\Theta} p(y\mid x, \DD,
    \FF_\Theta)p(\FF_\Theta)\mathrm{d}\FF_\Theta,
\end{equation}
where $p(\FF_\Theta)$ is some unknown distribution over the space of
all possible parametric model families over all possible parameter
spaces. In practice, when deciding on a parametric model family, a
typical approach is to focus on a finite subset of possible modelling
approaches informed by {\it a priori} knowledge of the problem domain
and then choose the best family or families via a model selection
procedure (see, \eg, \cite{murphy2012machine}). Another approach is to
use Monte Carlo dropout~\cite{gal2016dropout} which ensembles models
obtained via edge dropout (see Appendix~\ref{sec:dropout}). Figure~\ref{fig:combinations}, plot
$\FF_\theta$ in box (a), pictorially shows how selecting different
models (which in turns produces different loss landscapes) may lead to
different solutions.

\subsection{Multiple marginalisation}
Until now, we have focussed on marginalising out single
conditioning random variables to obtain better approximations to
$p(\theta\mid\DD)$. A natural step at this point is to combine
two or more of the proposed marginalisations.

In Figure~\ref{fig:combinations}, boxes (b)-(d), all the possible
combinations of marginalisations in our framework are depicted, some
of which have recently been studied. For example,
multiSWAG~\cite{wilson_bayesian_2020}, 
an extension to SWAG that can
be seen as an approximation to marginalising out $t$ and $\theta_0$ to
give
\begin{align*}
    p(\theta\mid \DD, h) =
    \int_{\theta_0}\lim_{t\to\infty}
    \frac{1}{t-\bar{\tau}}\sum_{\tau=\bar{\tau}}^t{\delta(\theta-A^t_h(\theta_0)\mid\DD)}p(\theta_0)\mathrm{d}\theta_0.
\end{align*}
This is depicted in Figure~\ref{fig:combinations}, box (b), plot
$t,\theta_0$. 
Another example is hyper-ensembles~\cite{wenzel2020hyperparameter} which can be seen as marginalising out both $\theta_0$ and the
hyperparameter space (Figure~\ref{fig:combinations}, box
(b), plot $\theta_0,A$). 

In general, this framework allows a number of marginalisation
combinations that have not been addressed in the literature thus far. Depending on the particular
problem at hand, different combinations may lead to improved
uncertainty estimates.

\section{Estimating uncertainty}
Until now, we have focussed on estimating uncertainty in the parameter
solution space. 
By using~\eqref{eq:prediction}, that
uncertainty can be propagated to the predictive distribution. 
\begin{table*}[t!]
  \tiny
  \centering
  
  \scalebox{0.8}{
      \begin{filecontents*}{results_NLL_400.csv}
Marginalisation,boston housing,concrete,energy,kin8nm,naval,power plant,protein,wine,yacht
$\MM_\theta$,$13.134 \pm 3.835$,$9.342 \pm 1.388$,$1.487 \pm 0.352$,$3.374 \pm 0.590$,$-4.395 \pm 3.015$,$32.769 \pm 4.371$,$96.558 \pm 13.491$,$23.328 \pm 7.372$,$1.300 \pm 0.543$
$t$,$5.958 \pm 1.962$,$5.752 \pm 1.092$,$1.278 \pm 0.164$,$0.198 \pm 0.551$,$-4.464 \pm 0.049$,$14.668 \pm 3.007$,$11.097 \pm 5.094$,$10.749 \pm 2.840$,$0.733 \pm 0.183$
$\theta_0$,$5.681 \pm 2.039$,$7.162 \pm 1.961$,$2.058 \pm 1.392$,$2.636 \pm 1.209$,\color{blue}$\mathbf{-6.020 \pm 0.173}$,$23.229 \pm 6.921$,$35.612 \pm 7.356$,$6.649 \pm 1.791$,$1.414 \pm 2.311$
$\MM_\theta$ + $t$,$4.096 \pm 0.730$,$4.365 \pm 0.810$,$1.310 \pm 0.087$,$-0.596 \pm 0.250$,$-4.417 \pm 0.065$,$9.824 \pm 1.965$,$11.526 \pm 3.242$,$7.121 \pm 1.806$,$0.763 \pm 0.106$
$\MM_\theta$ + $\theta_0$,$4.015 \pm 1.287$,$4.351 \pm 0.417$,\color{blue}$\mathbf{1.032 \pm 0.136}$,$-0.210 \pm 0.189$,$-5.847 \pm 0.095$,$12.105 \pm 1.754$,$24.517 \pm 1.083$,$4.157 \pm 0.858$,\color{blue}$\mathbf{0.525 \pm 0.198}$
$t$ + $\theta_0$,$\mathbf{3.511 \pm 0.816}$,$3.930 \pm 0.387$,$1.230 \pm 0.086$,$-0.673 \pm 0.204$,$-4.449 \pm 0.024$,$11.539 \pm 1.688$,$\mathbf{8.910 \pm 2.600}$,$4.022 \pm 0.772$,$\mathbf{0.634 \pm 0.147}$
$\MM_\theta$ + $t$ + $\theta_0$,\color{blue}$\mathbf{2.998 \pm 0.593}$,\color{blue}$\mathbf{3.494 \pm 0.286}$,$1.317 \pm 0.055$,\color{blue}$\mathbf{-0.977 \pm 0.123}$,$-4.425 \pm 0.023$,\color{blue}$\mathbf{8.223 \pm 1.159}$,\color{blue}$\mathbf{7.030 \pm 2.080}$,\color{blue}$\mathbf{3.035 \pm 0.559}$,$0.741 \pm 0.133$
\end{filecontents*}
\pgfplotstabletypeset[
          col sep=comma,
          string type,
          every column/.style={column type=r},
          every head row/.style={before row=\hline,after row=\hline},
          every last row/.style={after row=\hline},
      ]{results_NLL_400.csv}
  }
  \caption{NLL on UCI data sets expressed
    and mean $\pm$ standard deviation over the 20 train-test folds
    (5 for the protein dataset). Blue represents the best values and
    bold those whose mean is within the error bound of the
    best. Note that marginalising out $\MM_\theta$, $t$ and
    $\theta_0$ is approximately by MC-dropout, SWAG and ensembles
    respectively.}
  \label{tab:UCI_NLL}
\end{table*}
However, computing it explicitly is often intractable and approximations are preferred.
Monte Carlo sampling is
arguably the most widely used approach to approximating $p(y\mid x,
\DD, \FF_\Theta, *)$ as
\begin{equation*}
    \hat{p}(y\mid x, \DD, \FF_\Theta, *) = \frac{1}{K}\sum_{k=1}^K
    p(y\mid x,\DD,\theta_k,*),
\end{equation*}
where $K$ is the number of samples drawn from the trained model
parameter space and $\theta_k\sim \hat{p}(\theta\mid\DD,*)$. As $K$
increases, the accuracy of $\hat{p}(y\mid x, \DD, \FF_\Theta, *)$ also
increases. Similarly, the mean and variance
of the predictive distribution can be approximated as
$\hat{\mu}_{\hat{Y}}= \frac{1}{K}\sum_{k=1}^Kf_{\theta_k}(x)$ and $\hat{\sigma}_{\hat{Y}}^2=\frac{1}{K}\sum_{k=1}^K (f_{\theta_k}(x)-\mu_{\hat{Y}})^2$, 
with $\theta_k\sim \hat{p}(\theta\mid\DD,*)$.

Assumed Density Filtering (ADF)~\cite{ghosh2016assumed} 
can also be used to perform to
compute the expected value and variance of $p(y\mid x, \DD,
\FF_\Theta, *)$ in a single forward pass (see Appendix~\ref{sec:one_shot} for further details).

\section{Experimental results}
In this section, we apply methods described by the proposed framework on benchmarking
problems to investigate which forms and combinations of marginalisation 
are most useful from a practical point of view. We consider problems on benchmark UCI datasets and CIFAR-10. 
Additional details and experiments are provided in Appendix~\ref{appendix:results}.

\subsection{Regression on UCI datasets}\label{sec:UCI}
We consider regression problems on UCI data sets,
originally proposed in~\cite{hernandez2015probabilistic} and used for
benchmarking MC-dropout~\cite{gal2016dropout},
SWAG~\cite{maddox_simple_2019} and deep
ensembles~\cite{blundell2015weight}.  Each data set is split into 20
train-test folds except for \emph{protein} where 5 folds were used.
The network architecture had one hidden layer consisting of $50$ units
($100$ for the \emph{protein} dataset), followed by a dropout layer
with dropout rate fixed at $0.01$.  Each model was trained for $400$
epochs
to minimise the Mean Squared Error with no regularisation, using
Adam with a learning rate $\alpha=0.01$ and batch-size $b=100$.

We used SWAG, MC-dropout and $5$-model ensembles to approximate the
marginalisation of $t$, $\FF_\theta$ (as shown in the appendix) and $\theta_0$ respectively.
Results on all marginalisation combinations are reported in
Table~\ref{tab:UCI_NLL} in terms of negative log-likelihood (NLL).
The tendency is for NLL to decrease (and the
quality of the uncertainty estimate to improve) as more random
variables are marginalised out, with combined $\FF_\theta$, $t$ and
$\theta_0$ producing the lowest NLL in 6 of 9 data sets.

\subsection{Classification on CIFAR-10}
We train VGG16 neural networks for $300$ epochs over CIFAR-10 using SGD with
batch-size $b=128$. 
We approximated the marginalisation over
$t$ and $\MM_\theta$ using SWAG and MC-dropout respectively, while
ensembles were used to marginalise over $\theta_0$ and learning rates
$\alpha$ (see Appendix~\ref{sec:app_cifar} for furhter detais).

\begin{figure}[h!]
    \centering
    \begin{subfigure}{0.4\linewidth}
        \centering
        \includegraphics[width=\linewidth]{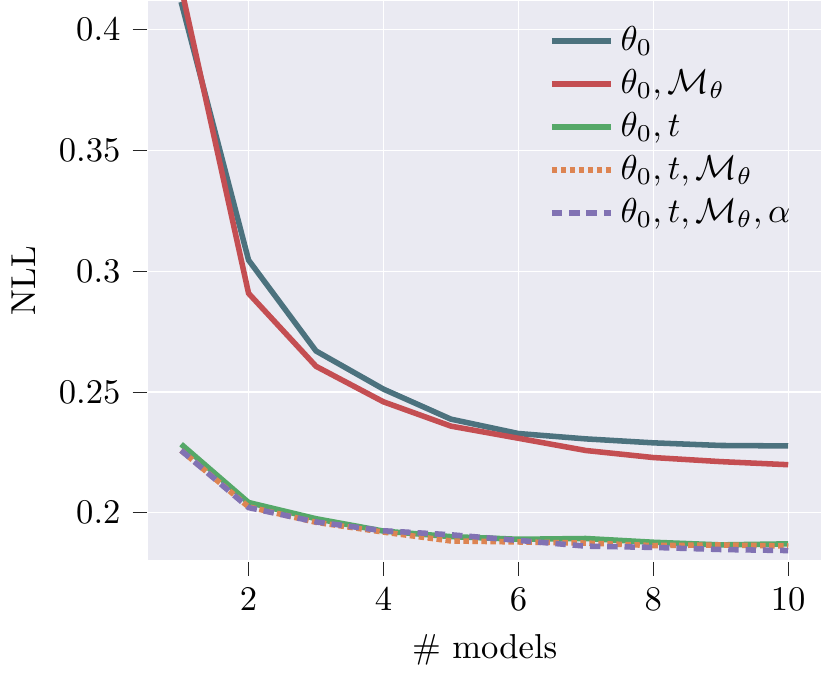}  
    \end{subfigure}
    \begin{subfigure}{0.4\linewidth}
        \centering
        \includegraphics[width=\linewidth]{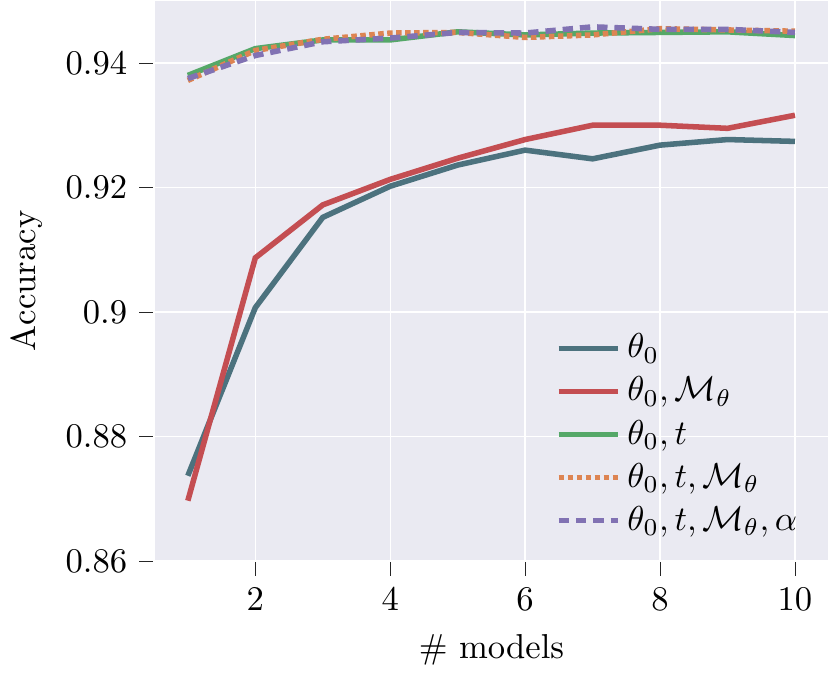}   
    \end{subfigure}
    \caption{Evolution of NLL and accuracy over the CIFAR dataset.}
    \label{fig:cifar}
\end{figure}

The NLL and accuracy on the test data set with increasing ensemble size
are reported in Figure~\ref{fig:cifar}. As we increase the ensemble
size, the NLL decreases and the accuracy increases, an outcome we know
from~\cite{blundell2015weight}. However, one must note that even with a single model,
marginalising out additional random variables $t$, $\MM_\theta$ and
$\alpha$ produces a significantly lower NLL (and higher accuracy). 
In particular, while $t$ have a significant impact, other marginalisations seems to be less effective.
This trend continues when we add more models to the ensemble. 

\section{Discussion}
From the preliminary experiments we performed some trends seem to appear.
First, we see that marginalising $\theta_0$ and $t$ always produce good results.
This is probably due to the fact that these two marginalisations better capture 
the geometry of the parameter space for a fixed model and set of hyperparameters. 
Marginalising out other hyperparameters can be useful~\cite{wenzel2020hyperparameter}, 
but it probably requires some preliminary fine tuning to select a good distribution from which to draw them.
A similar reasoning can be applied to the marginalisation of the model family, where a preliminary search for 
good candidates seems to be required. MC-dropout with a very small dropout rate can be a good option to 
minimally perturb a good architecture.
If fine tuning is not an option, then marginalising $\theta_0$ and $t$ seems 
to be the best way to go among the options available in this framework.

\bibliography{biblio.bib}
\bibliographystyle{icml2021}

\appendix
\section{Details and additional experiments}\label{appendix:results}
\begin{figure*}[t!]
    \centering
    \includegraphics[width=\linewidth]{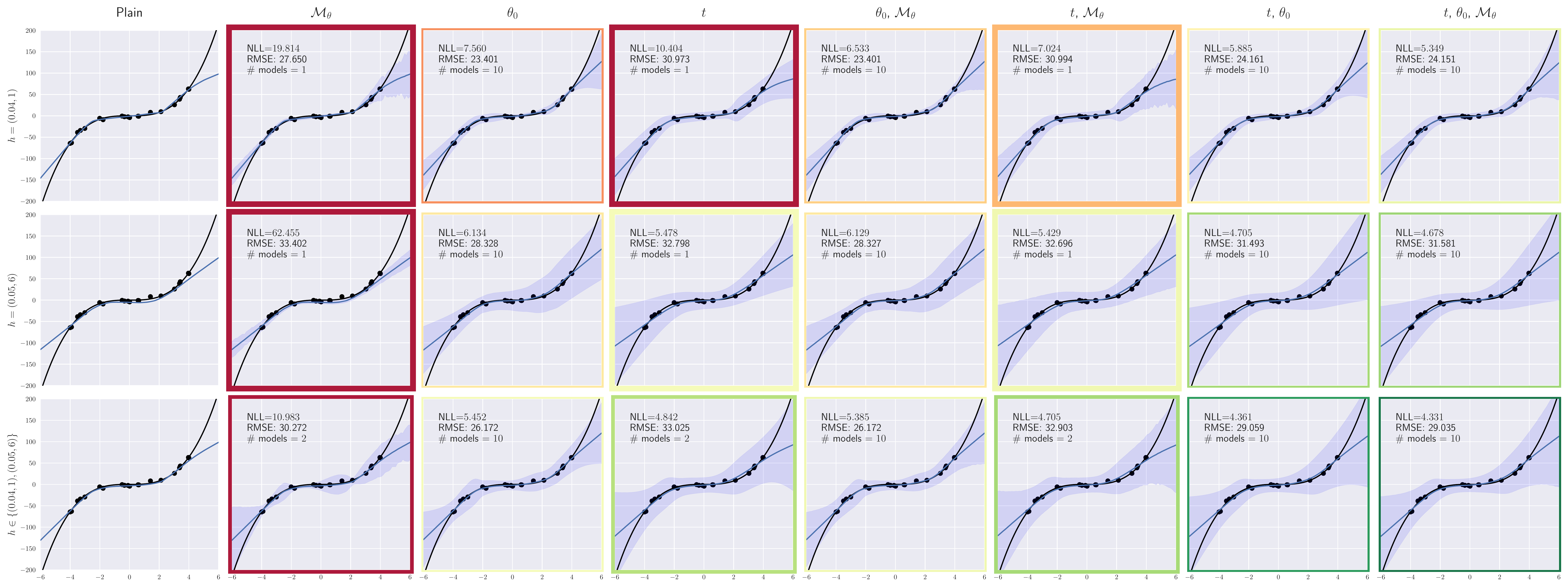}  
    \caption{Regression on toy data. Training samples are depicted
        as blue dots. The black line denotes the true data generating
        function, while the blue line denotes the predictive mean of
        the trained model. The shadowed area denotes the $3\sigma$
        confidence interval. Each plot (except for those in the first
        column) has a border that is coloured according to the NLL on
        test data, with green indicating low NLL and red high. The
        thickness of the border represents the number of trained
        models in the ensemble: the thicker the border, the smaller
        the ensemble. The NLL, RMSE and ensemble size are also
        reported in each plot.}
    \label{fig:toy}
\end{figure*}
\subsection{Regression on toy data}
We first consider a simple 1-dimensional toy regression dataset
consisting of $10$ training samples generated from $y=x^3+\epsilon$
with $\epsilon\sim\NN(0,9)$ and $x\in \UU[-4,4]$, as
in~\cite{hernandez2015probabilistic,blundell2015weight,franchi2019tradi},
to evaluate the performance of different marginalisations. A neural
network with one hidden layer consisting of $100$ units with ReLU
activation was trained for $100$ epochs to fit the data using the
stochastic gradient descent algorithm.
We compared the effects of approximating the marginalisation over $t$
and $\theta_0$ with hyperparameters
$h=(\alpha,b)\in\{(0.04,1),(0.05,6)\}$. The marginalisation over $t$
was approximated using SWAG, while over $\theta_0$, we used an
ensemble of $20$ models. RMSE and NLL were calculated on a test set of
1000 equally spaced points $(x,x^3)$, $x\in[-6,-6]$ that lie on the
black line.

The results for the various marginalisation strategies over the
different hyperparameter combinations are depicted in
Figure~\ref{fig:toy}. The first and second rows correspond to models
trained with hyperparameter points $h=(0.04,1)$ and $h=(0.05,6)$
respectively, while results in the third row are obtained by ensembles
combining the two hyperparameter point. For the first, third and fifth
columns, we ensembled over two models (one for each hyperparameter
point), whereas the remaining columns, corresponding to marginalising
out $\theta_0$, are ensembles of 5 initial conditions, trained once
with $h=(0.04,1)$ and once with $h=(0.05,6)$.

Our results show that the quality of predictive uncertainty clearly
benefits from multiple marginalisations with a decreasing NLL trend
along the rows and columns. Interesting points to note are that
despite the hyperparameter choice, $h=(0.05,6)$ often produces a lower
NLL than $h=(0.04,1)$ and that the combined hyperparameters (third
row) outperforms both individual hyperparameters (first and second
rows). For RMSE, $h=(0.04,1)$ outperforms $h=(0.05,6)$, with the two
combined generally lying somewhere in between, but with the additional
advantage of a lower NLL. Also worthy of note is the computation time,
measured in terms of number of trained models. While marginalising out
$\theta_0$ will outperform other random variable combinations given a
sufficiently large ensemble, the point is that good estimates can be
achieved with limited computational budget, \eg, combined
hyperparameters with SWAG and MD-dropout (third row, sixth column).
\begin{table}[h!] 
    \tiny
    \centering
\begin{filecontents*}{datasets_details.csv}
Dataset,N,n,m,a
boston housing,506,13,1,0.050
concrete,1030, 8,1, 0.015
energy,768,8,1 (2), 0.056
kin8nm,8192,8,1,0.017
naval,11934,16,1 (2), 0.033
power plant,9568,4,1,0.099
protein,45730,9,1,0.099
wine,1599,11,1,0.013
yacht,308,6,1,0.099
\end{filecontents*}
\pgfplotstabletypeset[
    col sep=comma,
    string type,
    columns/Dataset/.style={fixed,fixed zerofill,precision=0,column type=l},
    columns/a/.style={column name={$\bar{\alpha}$}},
    columns/b/.style={column name={$\bar{b}$}},
    every column/.style={column type=r},
    every head row/.style={before row=\hline,after row=\hline},
    every last row/.style={after row=\hline},
    ]{datasets_details.csv}
\caption{Details on UCI data sets.}
\label{tab:UCI_data}
\end{table}
\begin{table*}[h!]
    \tiny
    \centering
    \scalebox{0.8}{
\begin{filecontents*}{results_RMSE_40.csv}
Marginalisation,boston housing,concrete,energy,kin8nm,naval,power plant,protein,wine,yacht
$\MM_\theta$,$\mathbf{3.247 \pm 0.622}$,$6.584 \pm 0.957$,$1.806 \pm 0.541$,$0.118 \pm 0.009$,$0.002 \pm 0.001$,$4.818 \pm 0.564$,$5.140 \pm 0.129$,$\mathbf{0.660 \pm 0.043}$,$1.387 \pm 0.448$
$t$,$\mathbf{2.900 \pm 0.600}$,$\mathbf{5.535 \pm 0.451}$,$1.570 \pm 0.313$,$\mathbf{0.100 \pm 0.004}$,$0.003 \pm 0.000$,$\mathbf{4.289 \pm 0.199}$,$5.072 \pm 0.159$,$\mathbf{0.633 \pm 0.041}$,$\mathbf{1.134 \pm 0.254}$
$\theta_0$,\color{blue}$\mathbf{2.798 \pm 0.660}$,$\mathbf{5.467 \pm 0.424}$,$\mathbf{1.278 \pm 0.115}$,$0.102 \pm 0.004$,$\mathbf{0.002 \pm 0.000}$,$\mathbf{4.341 \pm 0.270}$,\color{blue}$\mathbf{4.929 \pm 0.065}$,$\mathbf{0.635 \pm 0.040}$,\color{blue}$\mathbf{0.890 \pm 0.278}$
$\MM_\theta$ + $t$,$\mathbf{2.954 \pm 0.643}$,$\mathbf{5.597 \pm 0.446}$,$1.568 \pm 0.276$,$\mathbf{0.100 \pm 0.005}$,$0.003 \pm 0.001$,$\mathbf{4.276 \pm 0.182}$,$\mathbf{4.932 \pm 0.054}$,$\mathbf{0.634 \pm 0.042}$,$\mathbf{1.107 \pm 0.252}$
$\MM_\theta$ + $\theta_0$,$\mathbf{2.834 \pm 0.617}$,\color{blue}$\mathbf{5.445 \pm 0.429}$,\color{blue}$\mathbf{1.221 \pm 0.171}$,$0.103 \pm 0.004$,\color{blue}$\mathbf{0.002 \pm 0.000}$,$\mathbf{4.331 \pm 0.160}$,$\mathbf{4.943 \pm 0.060}$,$\mathbf{0.634 \pm 0.041}$,$\mathbf{0.913 \pm 0.295}$
$t$ + $\theta_0$,$\mathbf{2.917 \pm 0.678}$,$\mathbf{5.510 \pm 0.407}$,$1.475 \pm 0.145$,\color{blue}$\mathbf{0.098 \pm 0.003}$,$0.003 \pm 0.000$,$\mathbf{4.261 \pm 0.185}$,$\mathbf{4.940 \pm 0.066}$,$\mathbf{0.631 \pm 0.039}$,$\mathbf{1.024 \pm 0.220}$
$\MM_\theta$ + $t$ + $\theta_0$,$\mathbf{2.912 \pm 0.660}$,$\mathbf{5.464 \pm 0.372}$,$1.481 \pm 0.158$,$\mathbf{0.098 \pm 0.003}$,$0.003 \pm 0.000$,\color{blue}$\mathbf{4.258 \pm 0.179}$,$\mathbf{4.944 \pm 0.073}$,\color{blue}$\mathbf{0.630 \pm 0.039}$,$\mathbf{1.006 \pm 0.231}$
\end{filecontents*}
\pgfplotstabletypeset[
    col sep=comma,
    string type,
    every column/.style={column type=r},
    every head row/.style={before row=\hline,after row=\hline},
    every last row/.style={after row=\hline},
    ]{results_RMSE_40.csv}
    }

    \vspace{2ex}

    \scalebox{0.8}{
\begin{filecontents*}{results_NLL_40.csv}
Marginalisation,boston housing,concrete,energy,kin8nm,naval,power plant,protein,wine,yacht
$\MM_\theta$,$51.813 \pm 42.183$,$\sim10^{2}$,$\sim10^{2}$,$21.343 \pm 19.598$,$2.788 \pm 18.344$,$26.959 \pm 13.069$,$\sim10^{6}$,$\sim10^{4}$,$9.075 \pm 13.178$
$t$,$3.360 \pm 0.443$,$3.764 \pm 0.246$,$\mathbf{2.129 \pm 0.178}$,$-0.615 \pm 0.100$,$-3.812 \pm 0.111$,$3.305 \pm 0.167$,$7.725 \pm 3.579$,$8.861 \pm 2.216$,$1.462 \pm 0.181$
$\theta_0$,$6.471 \pm 3.983$,$5.692 \pm 2.593$,$2.651 \pm 1.768$,$0.972 \pm 0.747$,$\mathbf{-4.759 \pm 0.152}$,$6.490 \pm 2.578$,$21.574 \pm 10.052$,$16.994 \pm 3.050$,$\mathbf{1.219 \pm 0.324}$
$\MM_\theta$ + $t$,$\mathbf{3.321 \pm 0.735}$,$3.684 \pm 0.241$,$\mathbf{2.170 \pm 0.166}$,$-0.713 \pm 0.092$,$-3.776 \pm 0.252$,$3.252 \pm 0.130$,$7.542 \pm 1.405$,$9.034 \pm 2.350$,$1.434 \pm 0.142$
$\MM_\theta$ + $\theta_0$,$4.738 \pm 1.919$,$4.779 \pm 0.739$,\color{blue}$\mathbf{2.097 \pm 0.389}$,$0.037 \pm 0.240$,\color{blue}$\mathbf{-4.858 \pm 0.117}$,$5.062 \pm 1.122$,$16.541 \pm 2.250$,$17.652 \pm 6.301$,\color{blue}$\mathbf{1.095 \pm 0.336}$
$t$ + $\theta_0$,$\mathbf{2.847 \pm 0.426}$,$\mathbf{3.492 \pm 0.134}$,$\mathbf{2.114 \pm 0.097}$,$\mathbf{-0.794 \pm 0.054}$,$-3.837 \pm 0.044$,$\mathbf{3.138 \pm 0.127}$,$5.165 \pm 1.392$,$\mathbf{7.390 \pm 1.010}$,$\mathbf{1.415 \pm 0.131}$
$\MM_\theta$ + $t$ + $\theta_0$,\color{blue}$\mathbf{2.842 \pm 0.500}$,\color{blue}$\mathbf{3.450 \pm 0.132}$,$\mathbf{2.146 \pm 0.117}$,\color{blue}$\mathbf{-0.822 \pm 0.058}$,$-3.786 \pm 0.100$,\color{blue}$\mathbf{3.114 \pm 0.134}$,\color{blue}$\mathbf{3.677 \pm 0.488}$,\color{blue}$\mathbf{6.833 \pm 1.452}$,$1.456 \pm 0.122$
\end{filecontents*}
\pgfplotstabletypeset[
    col sep=comma,
    string type,
    every column/.style={column type=r},
    every head row/.style={before row=\hline,after row=\hline},
    every last row/.style={after row=\hline},
    ]{results_NLL_40.csv}
    }
\caption{Training for $40$ epochs with $\alpha=0.1$. RMSE (top) and
  NLL (bottom) on UCI data sets expressed and mean $\pm$ standard
  deviation over the 20 train-test folds (5 for the protein
  dataset). Blue represents the best metrics and bold those whose mean
  is within the error bound of the best.}
\label{tab:UCI_40epochs}
\end{table*}

\begin{table*}[h!]
    \tiny
    \centering
    \scalebox{0.8}{
\begin{filecontents*}{results_RMSE_hyp.csv}
Marginalisation,boston housing,concrete,energy,kin8nm,naval,power plant,protein,wine,yacht
$t$,$\mathbf{3.294 \pm 0.796}$,$\mathbf{5.386 \pm 0.431}$,$1.244 \pm 0.164$,$0.079 \pm 0.002$,$0.002 \pm 0.000$,$\mathbf{4.260 \pm 0.184}$,$4.894 \pm 0.041$,$\mathbf{0.623 \pm 0.044}$,$\mathbf{1.103 \pm 0.247}$
$\theta_0$,\color{blue}$\mathbf{3.125 \pm 0.624}$,$\mathbf{5.048 \pm 0.367}$,\color{blue}$\mathbf{0.573 \pm 0.086}$,\color{blue}$\mathbf{0.075 \pm 0.002}$,$\mathbf{0.001 \pm 0.000}$,$\mathbf{4.255 \pm 0.158}$,$4.880 \pm 0.052$,$\mathbf{0.615 \pm 0.038}$,\color{blue}$\mathbf{0.856 \pm 0.288}$
$\theta_0$ + $\alpha$,$\mathbf{3.141 \pm 0.740}$,\color{blue}$\mathbf{5.042 \pm 0.350}$,$\mathbf{0.595 \pm 0.092}$,$\mathbf{0.075 \pm 0.002}$,\color{blue}$\mathbf{0.001 \pm 0.000}$,$\mathbf{4.245 \pm 0.154}$,$\mathbf{4.874 \pm 0.081}$,\color{blue}$\mathbf{0.614 \pm 0.038}$,$\mathbf{0.871 \pm 0.280}$
$t$ + $\theta_0$,$\mathbf{3.231 \pm 0.778}$,$\mathbf{5.372 \pm 0.412}$,$1.111 \pm 0.114$,$\mathbf{0.075 \pm 0.002}$,$\mathbf{0.001 \pm 0.000}$,$\mathbf{4.225 \pm 0.158}$,$\mathbf{4.864 \pm 0.056}$,$\mathbf{0.619 \pm 0.042}$,$\mathbf{0.987 \pm 0.199}$
$t$ + $\theta_0$ + $\alpha$,$\mathbf{3.232 \pm 0.788}$,$\mathbf{5.372 \pm 0.408}$,$1.120 \pm 0.138$,$\mathbf{0.075 \pm 0.002}$,$\mathbf{0.001 \pm 0.000}$,\color{blue}$\mathbf{4.224 \pm 0.159}$,\color{blue}$\mathbf{4.840 \pm 0.040}$,$\mathbf{0.618 \pm 0.042}$,$\mathbf{0.981 \pm 0.190}$
\end{filecontents*}
\pgfplotstabletypeset[
    col sep=comma,
    string type,
    every column/.style={column type=r},
    every head row/.style={before row=\hline,after row=\hline},
    every last row/.style={after row=\hline},
    ]{results_RMSE_hyp.csv}
    }

    \vspace{2ex}

    \scalebox{0.8}{
\begin{filecontents*}{results_NLL_hyp.csv}
Marginalisation,boston housing,concrete,energy,kin8nm,naval,power plant,protein,wine,yacht
$t$,$3.548 \pm 0.847$,$\mathbf{3.426 \pm 0.208}$,$2.063 \pm 0.122$,$-1.100 \pm 0.040$,$-3.715 \pm 0.101$,$3.245 \pm 0.241$,$10.192 \pm 1.791$,$8.290 \pm 1.010$,$\mathbf{1.408 \pm 0.098}$
$\theta_0$,$6.332 \pm 2.875$,$7.386 \pm 2.497$,$\mathbf{1.306 \pm 0.341}$,$0.552 \pm 0.528$,$\mathbf{-5.129 \pm 0.258}$,$7.034 \pm 2.323$,$19.269 \pm 4.607$,$13.956 \pm 3.587$,\color{blue}$\mathbf{1.276 \pm 0.603}$
$\theta_0$ + $\alpha$,$5.663 \pm 2.183$,$7.322 \pm 2.172$,\color{blue}$\mathbf{1.215 \pm 0.390}$,$0.427 \pm 0.433$,\color{blue}$\mathbf{-5.261 \pm 0.173}$,$7.387 \pm 3.239$,$21.291 \pm 4.050$,$15.283 \pm 4.751$,$\mathbf{1.330 \pm 0.739}$
$t$ + $\theta_0$,\color{blue}$\mathbf{2.836 \pm 0.398}$,\color{blue}$\mathbf{3.269 \pm 0.159}$,$2.067 \pm 0.055$,$\mathbf{-1.136 \pm 0.022}$,$-3.699 \pm 0.039$,$\mathbf{3.017 \pm 0.056}$,\color{blue}$\mathbf{4.847 \pm 1.249}$,\color{blue}$\mathbf{5.726 \pm 0.795}$,$\mathbf{1.412 \pm 0.088}$
$t$ + $\theta_0$ + $\alpha$,$\mathbf{2.853 \pm 0.445}$,$\mathbf{3.274 \pm 0.162}$,$2.068 \pm 0.056$,\color{blue}$\mathbf{-1.139 \pm 0.022}$,$-3.704 \pm 0.038$,\color{blue}$\mathbf{3.014 \pm 0.074}$,$\mathbf{5.522 \pm 1.850}$,$\mathbf{5.728 \pm 0.792}$,$\mathbf{1.410 \pm 0.089}$
\end{filecontents*}
\pgfplotstabletypeset[
    col sep=comma,
    string type,
    every column/.style={column type=r},
    every head row/.style={before row=\hline,after row=\hline},
    every last row/.style={after row=\hline},
    ]{results_NLL_hyp.csv}
    }
\caption{Training for $40$ epochs with tuned learning rates. RMSE
  (top) and NLL (bottom) on UCI data sets expressed and mean $\pm$
  standard deviation over the $20$ train-test folds ($5$ for the protein
  dataset). Blue represents the best metrics and bold those whose mean
  is within the error bound of the best.}
\label{tab:UCI_40epochs_lr}
\end{table*}
\subsection{UCI datasets}
\subsubsection{Details on  datasets} 
Recall that a given data set $\DD=\{x_i,y_i\}_{i=1}^N$, with
$x_i\in\R^n$, $y_i\in\R^m$, has number of samples $N$, (input) random
variable dimension $n$ and (output) random variable dimension $m$.
Table~\ref{tab:UCI_data} provides additional details on the UCI data sets used for our experiments and the tuned learning rates
$\bar{\alpha}$ we will use in the next section. As in~\cite{gal2016dropout} and other works using these UCI data sets, for \emph{energy} and \emph{naval}, we
predict the first output variable so, $m=1$.

\subsubsection{Training for 40 epochs}
Here we report the results obtained by training the networks for $40$
epochs with two different set-ups: firstly using Adam with a fixed
learning rate $\alpha=0.1$ for all data sets and secondly using the
tuned learning rates $\alpha=\bar{\alpha}$ reported in
Table~\ref{tab:UCI_data}.

\paragraph{Fixed learning rate}
The results for the first case are reported in
Table~\ref{tab:UCI_40epochs}. Clearly, the number of epochs is too low
for MC-dropout to converge (see~\cite{gal2016dropout}), thus the NLL
for $\MM_\theta$ is generally poor. However, when also marginalising
out $t$ and/or $\theta_0$ (again approximated via SWAG and an ensemble
of $5$ models respectively), the addition of MC-dropout generally
results in lower NLL suggesting that even with few training epochs,
MC-dropout may help when marginalised out in combination with other
random variables.

\paragraph{Tuned learning rates}
Results with tuned learning rates are reported in
Table~\ref{tab:UCI_40epochs_lr}. We approximate jointly marginalising
out $\alpha$ and $\theta_0$ by drawing 5 $\alpha$ samples from
$\NN(\bar{\alpha},\bar{\alpha}/100)$, one for each $\theta_0$ in the
ensemble, and combine this marginalisation approximation with other
SWAG and multiSWAG as required.
While we propose that aggressive hyperparameter tuning can be avoided
with multitple marginalisations, we do see an improvement in NLL when
compared to Table~\ref{tab:UCI_NLL} and for a number of data sets the
NLL decreases further by perturbing the value of $\alpha$.

\paragraph{Comments}
By comparing the results in
Tables~\ref{tab:UCI_NLL},~\ref{tab:UCI_40epochs}, and~\ref{tab:UCI_40epochs_lr},
we make several observations.
\begin{enumerate}
\item The trend of NLL decreasing as the number of marginalisation combinations increases is maintained.
\item Training for $40$ epochs with a (fixed) higher or tuned learning
  rate generally produces comparable or even better results than those
  reported in Table~\ref{tab:UCI_NLL}. However, this comes with a cost
  of higher RMSE for most data sets which is likely due to
  non-convergence after training for $40$ epochs and hence higher
  prediction error. In general, the training set up very much depends
  on the desired use of a prediction. In domains such as disease
  diagnostics, understanding prediction certainty is more valuable
  than accuracy.
\item Fine-tuning the learning rates leads to an improvement in NLL
  across all data sets. However, the addition of MC-dropout can
  produce better results without the need for hyperparameter fine
  tuning.
\end{enumerate}

\subsection{Implementation details for CIFAR-10}\label{sec:app_cifar}
As in~\cite{maddox_simple_2019}, the learning rate was defined as a decreasing
schedule along epochs $e=1,\dots,n_e$ as
\begin{equation*}
    \alpha_e = \begin{cases}
        \alpha_u &\text{if } e< 0.5n_e\\
        \frac{\alpha_u-\alpha_l}{0.4n_e}e &\text{if } e\in[0.5n_e,0.9n_e]\\
        \alpha_l &\text{if } e>0.9n_e
    \end{cases}
\end{equation*}
for some $\alpha_u>\alpha_l$. When approximating the marginalisation over the learning rate $\alpha$, we randomly selected
$\alpha_u\in \UU[0.04,0.06]$ and $\alpha_l\in\UU[0.008, 0.012]$ when
marginalising out $\alpha$, otherwise we fixed $\alpha_u=0.05$ and
$\alpha_l=0.01$.

\section{Algorithm marginalisation}\label{sec:algos}
We first provide a detailed description of marginalising out
hyperparameters in the case of stochastic gradient descent. Then, we
extend this idea to the (countably infinite) set of optimisation
algorithms typically used in machine learning.
\paragraph{Stochastic gradient descent}
Over a single iteration, the SGD is governed by the step size
$\alpha\in\R$, the batch size $b\in[N]=\{1,\ldots,N\}$ and the batch
number $i\in[N_b]$, where $N_b=\ceil*{\frac{N}{b}}$, which without
loss of generality, corresponds to $$\BB_{b,i}=\{(x_{ib+1},
y_{ib+1}),\ldots,(x_{(i+1)b},y_{(i+1)b})\}\subset\DD.$$ In the context
of our framework, the hyperparameter space $H$ is defined as
\begin{align*}
  H=\big\{(\alpha,b,i)\in\R\times[N]\times[N]\mid i\leq N_b\big\},
\end{align*}
where points $p(\alpha,b,i)$ are drawn randomly with probability
$p(\alpha,i\mid b)\times p(b)$, $p(\alpha,i\mid
b)=p(\alpha)\times\UU\{1,N_b\}$ and $p(\alpha)$ and $p(b)$ are assumed
to be well chosen distributions, for example, through hyperparameter
optimisation. Note that $p(\alpha)$ is a continuous distribution
whereas $p(b)$ and $\UU\{1,N_b\}$ are discrete. Since random variables
$\mathrm{A}, B$ and $I$ are independent, taking products of their
respective marginal distributions is well defined.

To allow for scheduling of step size and batch size, over the course
of a complete training run, the hyperparameter space $H^t$ over which
we marginalise is defined as
\begin{align*}
  \big\{(\vect{\alpha,b,i})\in\R^t\times{[N]}^t\times{[N]}^t\mid
  i_\tau\leq N_b\big\},
\end{align*} 
where variable vectors are indexed over
$\tau\in\{1,\ldots,t\}$. Hyperparameter points are drawn randomly with
probability
\begin{align*}
  p(\vect{\alpha,b,i})=p(\vect{\alpha})\times\Pi_{\tau=1}^t\UU\{1,N_b\}\times
  p(\vect{b}).
\end{align*}
Note that at each iteration, we are compelled to select the next batch
for processing uniformly at random to ensure that, in the limit, the
gradient estimator is unbiased. If $t$ happens to be a multiple of
$N_b$, we can consider $t$ as a number of epochs, each of which
processes every batch exactly once. Thus the stochastic gradient is
guaranteed to be an unbiased estimator which allows us to select a
more appropriate distribution for batch ordering.

We can now marginalise out each of the hyperparameter variables to
express $p(\theta\mid \DD,t,\theta_0)$ as 
\begin{equation*}%\label{eq:ASGD}
  \sum_{\vect{b}\in{[N]}^t}\frac{1}{N_b^t}p(\vect{b})\sum_{\vect{i}\in[N_b]^t}{\int_{\vect{\alpha}}\delta(\theta-A^t_{\vect{\alpha,b,i}}(\theta_0)\mid\DD)p(\vect{\alpha})\mathrm{d}{\vect{\alpha}}},
\end{equation*}
where $(\vect{\alpha,b,i})\in H^t$.
\paragraph{The class of optimisation algorithms} 
Let $\AA=\{A_{h_j}\}_j$ be a (countably infinite) set of optimisation
algorithms $A_{h_j}$ that, given an initial condition $\theta_0$ and a
point in the appropriate hyperparameter space $h_j\in H_j$, return a
solution $A_{h_j}^t(\theta_0)$ to the given problem after $t$
iterations.
Let $\gamma_j\in\{0,1\}$ be a weight associated to the output of
algorithm $A_{h_j}$ and assume that
\begin{equation}\label{eq:gamma}
    \sum_j\gamma_j=1.
\end{equation}    
Then, if we define 
\begin{equation}\label{eq:multiA}
    \theta_t = \sum_{j} \gamma_j A_{h_j}^t(\theta_0),
\end{equation}
the output of the $j$-th algorithm is recovered by imposing
$\gamma_j=1$ (since, thanks to~\eqref{eq:gamma}, $\gamma_i=0$ for all
$i\neq j$).

Let $H\!=\!\prod_{j} H_j$ be the hyperparameter space associated to
$\AA$. We define $\Gamma\!\!=\!\!\left\lbrace\bm{\gamma}\in\prod_j
\mathbb{Z}_2 \mid\sum\gamma_j=1\right\rbrace$ and let
$\bm{A_{h}(\theta_0)}=[A_{h_j}^t(\theta_0)]_j\in\prod_j\R^d$. Then,
for some $\bm{\gamma}\in\Gamma$, we can rewrite~\eqref{eq:multiA} as
\begin{equation*}
    \theta_t = \langle (\bm{1}_d \otimes \bm{\gamma}),
    \bm{A_{h}(\theta_0)}\rangle.
\end{equation*}
Then, we marginalise out $\bm{\gamma}$ and $\bm{h}$ to express
$p(\theta\mid\DD,t,\theta_0)$ as
\begin{align*}
\int_{(\bm{\gamma},\bm{h})} \delta(\theta-\langle (\bm{1}_d \otimes \bm{\gamma}), \bm{A_{h}(\theta_0)}\rangle\mid\DD)p(\bm{\gamma},\bm{h})\mathrm{d}(\bm{\gamma},\bm{h}).
\end{align*}

\section{MC-dropout as model marginalisation}\label{sec:dropout}
Given a particular $\theta\in\Theta$ and parametric model family
$\FF_\Theta$, let $\MM_\theta\subset\FF_\Theta$ be the parametric
model family obtained by masking elements of $\theta$. This is a
finite set of models given by
\begin{align*}
  \MM_\theta =
  \{f_{\theta\odot\omega_1},f_{\theta\odot\omega_2},\ldots,f_{\theta\odot\omega_{2^d}}\},
\end{align*}
where the dropout masks $\omega_j, j=1,\ldots,d$ take values from
$\{0,1\}^d$ and $\odot$ is the Hadamard product. Masks $\omega$ can be
drawn randomly with probability $p(\omega)=\Pi_{j=1}^d
\mathbf{Bern}(p_j)$, $0\leq p_j\leq1$ and we can marginalise out
$\omega$ to give
\begin{equation*}
    p(y\mid x, \DD, \MM_\theta) =\sum_\omega p(y\mid x, \DD,\theta,
    \omega)p(\omega)
\end{equation*}
and then $\theta$ to give
\begin{align*}%\label{eq:mask}
  p(y\mid x,\DD,\FF_\Theta)&=\int_\theta p(y\mid x,\DD,\MM_\theta)\mathrm{d}\theta\nonumber\\
  &=\int_\theta\sum_\omega p(y\mid x,\DD,\theta,\omega)p(\omega)\mathrm{d}\theta\nonumber\\
  &=\int_\theta\sum_\omega p(y\mid x,\theta)p(\theta\mid\DD,\omega)p(\omega)\mathrm{d}\theta.
\end{align*}
Similar, we can condition on all the variables we have considered thus
far to express $p(y\mid x,\DD,\FF_\Theta,t,h,\theta_0)$ as
\begin{align*}
  \int_\theta\sum_\omega p(y\mid
  x,\theta)\underbrace{p(\theta\mid\DD,t,h,\theta_0,\omega)}_{\delta(\theta-A^t_h(\theta_0)\odot\omega\mid\DD)}p(\omega)\mathrm{d}\theta.
\end{align*}
The idea of marginalising out the entire class of parametric model
families can be formulated in a similar way to the algorithm
marginalisation strategy described in Section~\ref{sec:algos}.

\section{One-shot estimation of the predictive probability distribution}\label{sec:one_shot}
In the main paper we showed how to compute statistics of the predictive probability distribution via Monte Carlo sampling.
The main drawback of Monte Carlo sampling is the requirement to sample
multiple times from the parameter distribution in order to compute
$\hat{p}(y\mid x, \DD, \FF_\theta, *)$, which in turn, requires
multiple forward passes through the network.

To overcome this issue, Assumed Density Filtering (ADF) can be used to perform a one-shot estimation, where we
compute the expected value and variance of $p(y\mid x, \DD,
\FF_\Theta, *)$ in a single forward pass.
ADF has previously been used to propagate input uncertainty through
the network to provide output uncertainty~\cite{gast2018lightweight},
and to learn probability distributions over
parameters~\cite{ghosh2016assumed} (a similar approach has been used
in~\cite{wu2018deterministic}). Also in~\cite{brach2020single}, a
one-shot approach has been proposed to approximate uncertainty for
MC-dropout.

Let $Z^i = f_\theta^i(Z^{i-1})$ represent the output of
the $i$-th layer of a feed-forward neural network (considered as a
function between two random variables). We can convert the $i$-th
layer into an uncertainty propagation layer by simply matching first
and second-order central momenta~\cite{minka2001family}, \ie,
\begin{align}
    \mu_{Z^i} &= \E_{Z^{i-1}, \Theta^i}[f^i(Z^{i-1},\Theta^i)]\label{eq:os_mean}\\
    \sigma_{Z^i}^2 &= \Var_{Z^{i-1}, \Theta^i}[f^i(Z^{i-1},\Theta^i)]\label{eq:os_var}
\end{align}
where $Z^{i-1}\sim \NN(\mu_{Z^{i-1}}, \sigma_{Z^{i-1}})$ and
$\Theta^i\sim\NN(\mu_{\Theta^{i}}, \sigma_{\Theta^{i}})$. By doing so,
the values of $\hat{\mu}_{\hat{Y}}$ and $\hat{\sigma}_{\hat{Y}^2}$ are
obtained as the output of the last layer of the modified neural
network.
Notice that this procedure can account for input (aleatoric)
uncertainty as long as the input $Z^0\sim \NN(\mu_{Z^{0}},
\sigma_{Z^{0}})$. 

The main drawback with ADF is the reliance on a modification
to the structure of the neural network, where each layer is replaced
with its probabilistic equivalent. Also, while many probabilistic
layers admit a closed form solution to
compute~\eqref{eq:os_mean},~\eqref{eq:os_var} (see
below), in some cases approximation is
needed.

\subsection{Probabilistic layers}\label{appendix:one_shot}
\paragraph{Linear layer}
Let $X,W,B$ be independent random variables and consider a linear
layer, $f(x,w,b)= wx+b$. Then
\begin{align}
    \E[f(X)] &= \E[W]\E[X] + \E[B]\\
    \Var[f(X)] &= \Var[WX] + \Var[B]\nonumber\\
    &=\Var[W]\Var[X] + \Var[W]\E[X]^2 \nonumber\\
    &\hspace{3ex}+ \E[W]^2\Var[X] + \Var[B]
\end{align}
In this case, if $X,W,B$ are assumed to be normally distributed,
$f(x,W,b)$ can be approximated by gaussian with the above statistics.

\paragraph{ReLU layer}
Given a random variable $X\sim\NN(\mu,\sigma)$ the ReLU non-linearity
activation function $f(x)=\max\{0,x\}$ leads to a closed form solution
of its momenta~\cite{frey1999variational}.
\begin{align}
    \E[f(X)] &= \mu\Phi\left(\frac{\mu}{\sigma}\right) + \sigma\phi\left(\frac{\mu}{\sigma}\right)\\
    \Var[f(X)] &= (\mu^2+\sigma^2)\Phi\left(\frac{\mu}{\sigma}\right) + \mu\sigma\phi\left(\frac{\mu}{\sigma}\right) -  \E[f(X)]^2
\end{align}
where $\phi$ is the standard normal distribution function
\begin{equation*}
    \phi(z) = \frac{1}{\sqrt{2\pi}}e^{-\frac{1}{2}z^2}
\end{equation*}
and $\Phi$ is the cumulative normal distribution function
\begin{equation*}
    \Phi(z) = \frac{1}{\sqrt{2\pi}}\int_{-\infty}^z e^{-\frac{1}{2}t^2}\mathrm{d}t.
\end{equation*}

\paragraph{Other layers}
Almost all types of layers have closed form solutions or can be
approximated. See~\cite{Gast_2018_CVPR,loquercio2020uncertainty} for
more details and~\cite{brach2020single} for dropout layers.

\end{document}